\documentclass[runningheads]{llncs}
\usepackage{graphicx}
\usepackage{amsmath,amssymb} 
\usepackage{color}
\usepackage{floatrow}
\usepackage{multirow}
\usepackage{xspace}

\usepackage{float}
\floatstyle{plaintop}
\restylefloat{table}

\makeatletter
\DeclareRobustCommand\onedot{\futurelet\@let@token\@onedot}
\def\@onedot{\ifx\@let@token.\else.\null\fi\xspace}

\def\eg{\emph{e.g}\onedot} 
\def\ie{\emph{i.e}\onedot}

\def\etal{\emph{et al}\onedot}

\newcommand*\samethanks[1][\value{footnote}]{\footnotemark[#1]}

\makeatother

\begin{document}
\title{The Devil of Face Recognition is in the Noise} 

\titlerunning{The Devil of Face Recognition is in the Noise}
%
\author{Fei Wang \thanks{= equal contribution} \inst{1} \orcidID{0000-0002-1024-5867} \and
Liren Chen \samethanks \inst{2} \orcidID{0000-0003-0113-5233} \and \\
Cheng Li\inst{1} \orcidID{0000-0002-0892-4705} \and 
Shiyao Huang\inst{1} \orcidID{0000-0002-5198-2492} \and \\
Yanjie Chen\inst{1} \orcidID{0000-0003-1918-6776} \and
Chen Qian\inst{1} \orcidID{0000-0002-8761-5563} \and \\
Chen Change Loy \inst{3} \orcidID{0000-0001-5345-1591} 
}

%
\authorrunning{Fei Wang \etal}
%

\institute{SenseTime Research \and University of California San Diego \and Nanyang Technological University \\
\email{\{wangfei, chengli, huangshiyao, chenyanjie, qianchen\}@sensetime.com, lic002@eng.ucsd.edu, ccloy@ieee.org}}
\maketitle              
\begin{abstract}
The growing scale of face recognition datasets empowers us to train strong convolutional networks for face recognition. While a variety of architectures and loss functions have been devised, we still have a limited understanding of the source and consequence of label noise inherent in existing datasets. 
We make the following contributions: 
1) We contribute cleaned subsets of popular face databases, \ie, MegaFace and MS-Celeb-1M datasets, and build a new large-scale noise-controlled IMDb-Face dataset.
2) With the original datasets and cleaned subsets, we profile and analyze label noise properties of MegaFace and MS-Celeb-1M. We show that a few orders more samples are needed to achieve the same accuracy yielded by a clean subset.
3) We study the association between different types of noise, \ie, label flips and outliers, with the accuracy of face recognition models. 
4) We investigate ways to improve data cleanliness, including a comprehensive user study on the influence of data labeling strategies to annotation accuracy. 
The IMDb-Face dataset has been released on \url{https://github.com/fwang91/IMDb-Face}.

\end{abstract}

\setcounter{footnote}{0}
\section{Introduction}
\label{sec:introduction}
Datasets are pivotal to the development of face recognition. From the early FERET dataset~\cite{phillips1998feret} to the more recent LFW~\cite{huang2007labeled}, MegaFace~\cite{kemelmacher2016megaface,nech2017level}, and MS-Celeb-1M~\cite{guo2016ms}, face recognition datasets play a main role in driving the development of new techniques. The datasets not only become more diverse, the scale of data is also growing tremendously. For instance, MS-Celeb-1M~\cite{guo2016ms} contains around 10M images for 100K celebrities, far exceeding FERET~\cite{phillips1998feret} that only has 14,126 images from 1,199 individuals. Large-scale datasets together with the emergence of deep learning have led to the immense success of face recognition in recent years.

\begin{figure}[t]
\begin{center}
\includegraphics[width=\linewidth]{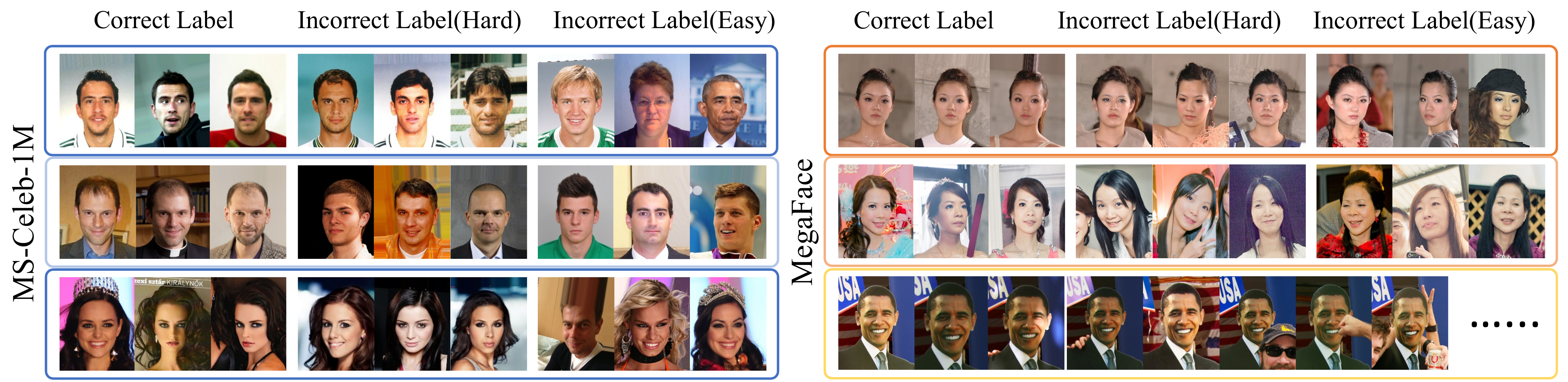}
\caption{Label noises in MegaFace~\cite{nech2017level} and MS-Celeb-1M~\cite{guo2016ms}. Each row depicts images that are labeled with the same identity. Some incorrect labels are easy while many of them are hard.}
\label{fig:overview}
\end{center}
\end{figure}

Large-scale datasets are inevitably affected by label noise. The problem is pervasive since well-annotated datasets in large-scale are prohibitively expensive and time-consuming to collect. That motivates researchers to resort to cheap but imperfect alternatives. 
A common method is to query celebrities' images by their names on search engines, and subsequently clean the labels with automatic or semi-automatic approaches~\cite{parkhi2015deep,li2016robust,deng2017marginal}.
Other methods introduce clustering with constraints on social photo sharing sites.
The aforementioned methods offer a viable way to scale the training samples conveniently but also bring label noises that adversely affect the training and performance of a model. 
We show some samples with label noises in Figure~\ref{fig:overview}. As can be seen, MegaFace~\cite{nech2017level} and MS-Celeb-1M~\cite{guo2016ms} consist considerable incorrect identity labels. Some noisy labels are easy to remove while many of them are hard to be cleaned. In MegaFace, there are a number of redundant images too (shown in the last row).

The first goal of this paper is to develop an understanding of the source of label noise and its consequences towards face recognition by deep convolutional neural networks (CNN)~\cite{sun2014deep,Schroff_2015_CVPR,wen2016discriminative,huang2018deep,cao2018pose,zhan2018consensus}. 
We seek answers to questions like: How many noisy samples are needed to achieve an effect tantamount to clean data? What is the relationship between noise and final performance? What is the best strategy to annotate face identities? 
A better understanding of the aforementioned questions would help us to design a better data collection and cleaning strategy, avoid pitfalls in training, and formulate stronger algorithms to cope with real-world problems. 
To facilitate our research, we manually clean subsets of two most popular face recognition databases, namely, MegaFace~\cite{nech2017level} and MS-Celeb-1M~\cite{guo2016ms}. We observe that a model trained with only $32\%$ of MegaFace or $20\%$ of MS-Celeb-1M cleaned subsets, can already achieve comparable performance with models that are trained on the respective full dataset. The experiments suggest that a few orders more samples are needed for face recognition model training if noisy samples are used. 

The second goal of our study is to build a clean face recognition dataset for the community. The dataset could help training better models and facilitate further understanding of the relationship between noise and face recognition performance. To this end, we build a clean dataset called \textbf{IMDb-Face}. The dataset consists of 1.7M images of 59K celebrities collected from movie screenshots and posters from the IMDb website\footnote{\url{www.IMDb.com}}. Due to the nature of the data source, the images exhibit large variations in scale, pose, lighting, and occlusion. We carefully clean the dataset and simulate corruption by injecting noise on the training labels. The experiments show that the accuracy of face recognition decreases rapidly and nonlinearly with the increase of label noises. 
In particular, we confirm the common belief that the performance of face recognition is more sensitive towards label flips (example has erroneously been given the label of another class within the dataset) than outliers (image does not belong to any of the classes under consideration, but mistakenly has one of their labels).
We also conduct an interesting experiment to analyze the reliability of different ways of annotating a face recognition dataset. We found that label accuracy correlates with time spent on annotation. The study helps us to find the source of erroneous labels and thereafter design better strategies to balance annotation cost and accuracy.
 
We hope that this paper could shed lights on the influences of data noise to the face recognition task, and point to potential labelling strategies to mitigate some of the problems. We contribute the new data \textbf{IMDb-Face} with the community. It could serve as a relatively clean data to facilitate future studies of noises in large-scale face recognition. It can also be used as a training data source to boost the performance of existing methods, as we will show in the experiments.

\section{How Noisy is Existing Data?}
\label{sec:datasets}

We first introduce some popular datasets used in face recognition study and then approximate their respective signal-to-noise ratio.

\subsection{Face Recognition Datasets}

Table~\ref{tab:valueExperiment} provides a summary of representative datasets used in face recognition research.

\begin{table}[t]
\begin{center}
\caption{Various face recognition datasets. }
\footnotesize
\resizebox{\linewidth}{!}{
\begin{tabular}{c | c | c | c | c | c}
  \hline
  Dataset & \#Identities & \#Images & Source & Cleaned? & Availablity  \\
  \hline
  LFW~\cite{huang2007labeled}     & 5K         & 13K    & Search Engine & Automatic Detection   & Public \\
  CelebFaces~\cite{sun2014deep,celebface} & 10K     & 202K   & Search Engine & Manually Cleaned      & Public  \\
  VGG-Face~\cite{parkhi2015deep}  &2.6K      &2.5M    & Search Engine & Semi-automated Clean  & Public  \\
  CASIA-WebFace~\cite{yi2014learning} &10k   &0.5M    & IMDb          & Automatic Clean       & Public  \\
  MS-Celeb-1M(v1)~\cite{guo2016ms} & 100k   & 10M    & Search Engine & None                  & Public \\
  MegaFace~\cite{nech2017level} & 670K      & 4.7M   & Flickr        & Automatic Cleaned     &Public \\
  Facebook~\cite{Taigman_2014_CVPR} & 4k        &4.4M    & --             & --                     & Private  \\
  Google~\cite{Schroff_2015_CVPR}    & 8M        &200M    & --             & --                     & Private  \\
  \textbf{IMDb-Face} & \textbf{59K}  & \textbf{1.7M}  & \textbf{IMDb} & \textbf{Manually Cleaned} & \textbf{Public} \\
  \hline
\end{tabular}
}
\end{center}
\label{tab:valueExperiment}
\end{table}

\noindent\textbf{LFW:} Labeled Faces in the Wild (LFW)~\cite{huang2007labeled} is perhaps the most popular dataset to date for benchmarking face recognition approaches. The database consists of $13,000$ facial images of $1,680$ celebrities. Images are collected from Yahoo News by running the Viola-Jones face detector. Limited by the detector, most of the faces in LFW is frontal. The dataset is considered sufficiently clean despite some incorrectly labeled matched pairs are reported. Errata of LFW are provided in \url{http://vis-www.cs.umass.edu/lfw/}.

\noindent\textbf{CelebFaces:} CelebFaces~\cite{sun2014deep,celebface} is one of the early face recognition training databases that are made publicly available. Its first version contains $5,436$ celebrities and $87,628$ images, and it was upgraded to $10,177$ identities and $202,599$ images in a year later. Images in CelebFaces were collected from search engines and manually cleaned by workers.

\noindent\textbf{VGG-Face:} VGG-Face~\cite{parkhi2015deep} contains 2,622 identities and 2.6M photos. More than 2,000 images per celebrity were downloaded from search engines. The authors treat the top 50 images as positive samples and train a linear SVM to select the top 1,000 faces. To avoid extensive manual annotation, the dataset was `block-wise' verified, \ie, ranked images of each identity are displayed in blocks and annotators are asked to validate blocks as a whole. In this study we did not focus on VGG-Face~\cite{parkhi2015deep} since it should have the similar `search-engine bias' problem with MS-Celeb-1M~\cite{guo2016ms}.

\noindent\textbf{CASIA-WebFace:} The images in CASIA-WebFace~\cite{yi2014learning} were collected from IMDb website. The dataset contains 500K photos of 10K celebrities and it is semi-automatically cleaned via tag-constrained similarity clustering. The authors start with each celebrity's main photo and those photos that contain only one face. Then faces are gradually added to the dataset constrained by feature similarity and name tag. CASIA-WebFace uses the same source as the proposed IMDb-Face dataset. However, limited by the feature and clustering steps, CASIA-WebFace may fail to recall many challenging faces. 

\noindent\textbf{MS-Celeb-1M:} MS-Celeb-1M~\cite{guo2016ms} contains 100K celebrities who are selected from the 1M celebrity list in terms of their popularities. Public search engines are then leveraged to provide approximately 100 images for each celebrity, resulting in about 10M web images.
The data is deliberately left uncleaned for several reasons. Specifically, collecting a dataset of this scale requires tremendous efforts in cleaning the dataset. Perhaps more importantly, leaving the data in this form encourages researchers to devise new learning methods that can naturally deal with the inherent noises.   

\noindent\textbf{MegaFace:}  Kemelmacher-Shlizerman \etal~\cite{nech2017level} clean massive number of images published on Flickr by proposing algorithms to cluster and filter face data from the YFCC100M dataset. For each user's albums, the authors merge face pairs with a distance closer than $\beta$ times of average distance. Clusters that contain more than three faces are kept. Then they drop `garbage' groups and clean potential outliers in each group.
A total of 672K identities and 4.7M images were collected. MegaFace2 avoids `search-engine' bias as in VGG-Face~\cite{parkhi2015deep} and MS-Celeb-1M~\cite{guo2016ms}. However, we found this cluster-based approach introduces new bias. MegaFace prefers small groups with highly duplicated images, \eg, face captured from the same video. Limited by the base model for clustering, considerable groups in MegaFace contain noises, or sometimes mess up multiple people in the same group.

\begin{figure}[t]
\begin{center}
   \includegraphics[width=0.7\linewidth]{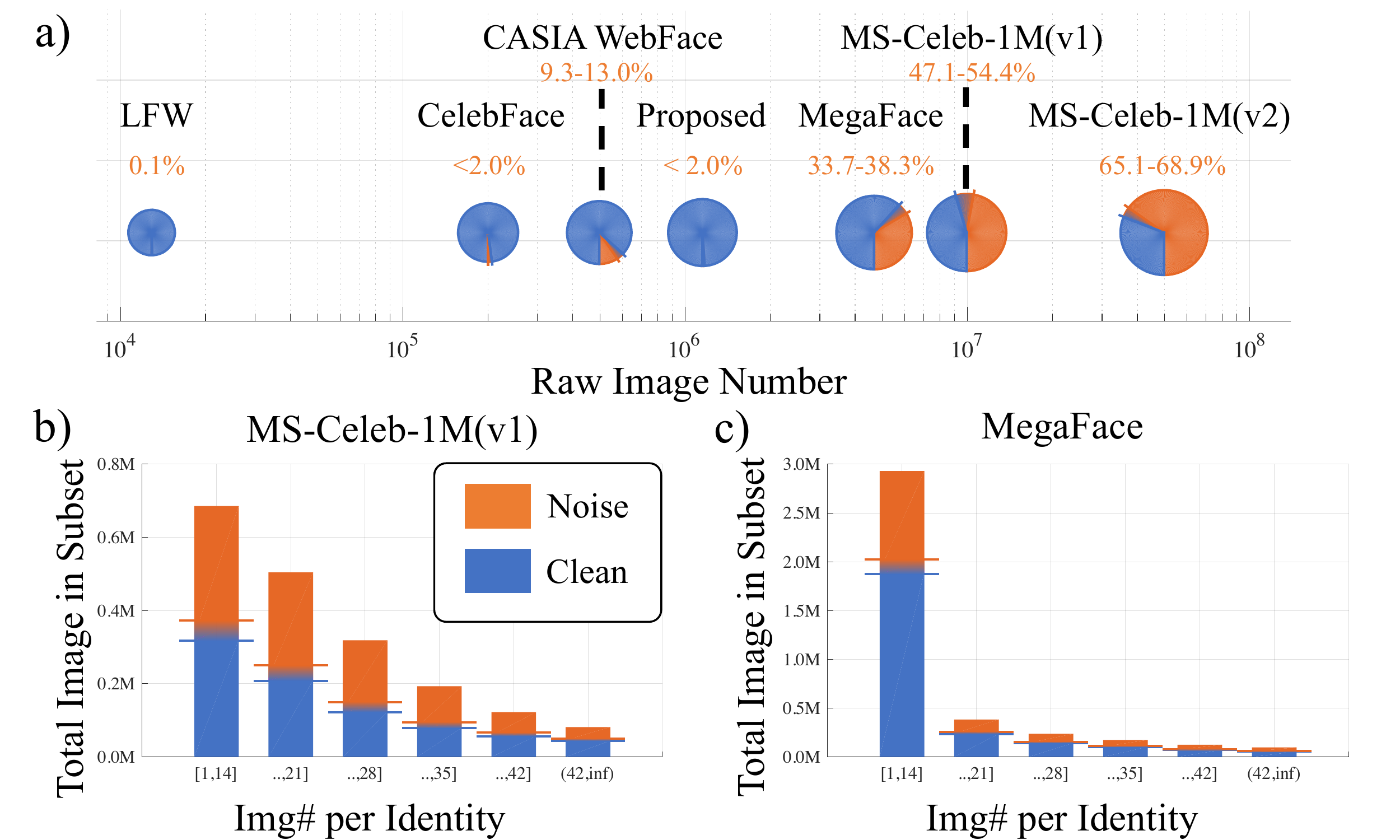}
\end{center}
   \caption{(a) A visualization of size and estimated noise percentage of datasets. (b) Noise distribution of MS-Celeb-1M(v1)~\cite{guo2016ms}. (c) Noise distribution of MegaFace~\cite{nech2017level}. The two horizontal lines in each bar represent the lower- and upper-bounds of noise, respectively. See Sec.~\ref{subsec:noise_ratio} for details.}
\label{fig:noiseDistribution}
\end{figure}

\subsection{An Approximation of Signal-to-Noise Ratio}
\label{subsec:noise_ratio}

Owing to the source of data and cleaning strategies, existing large-scale datasets invariably contain label noises. In this study, we aim to profile the noise distribution in existing datasets. Our analysis may provide a hint to future research on how one should exploit the distribution of these data. 

It is infeasible to obtain the exact number of these noises due to the scale of the datasets. We bypass this difficulty by randomly selecting a subset of a dataset and manually categorize them into three groups -- `correct identity assigned', `doubtful', and `wrong identity assigned'.
We select a subset of 2.7M images from MegaFace~\cite{nech2017level} and 3.7M images from MS-Celeb-1M~\cite{guo2016ms}. For CASIA-WebFace~\cite{yi2014learning} and CelebFaces~\cite{sun2014deep,celebface}, we sampled 30 identities to estimate their signal-to-noise ratio. The final statistics are visualized in Figure~\ref{fig:noiseDistribution}(a).
Due to the difficulty in estimating the exact ratio, we approximate an upper and a lower bound of noisy data during the estimation. The lower-bound is more optimistic considering doubtful labels as clean data. The upper-bound is more pessimistic considering all doubtful cases as badly labeled. We provide more details on the estimations in the supplementary material.
As observed in Figure~\ref{fig:noiseDistribution}(a), the noise percentage increases dramatically along the scale of data. This is not surprising given the difficulty in data annotation. 
It is noteworthy that the proposed IMDb-Face pushes the envelope of large-scale data with a very high signal-to-noise ratio (noise is under 10\% of the full data).

We investigate further the noise distribution of the two largest public datasets to date, MS-Celeb-1M~\cite{guo2016ms} and MegaFace~\cite{nech2017level}.
We first categorize identities in a dataset based on their number of images. A total of six groups/bins are established. We then plot a histogram showing the signal-to-noise ratio of each bin along the noise lower- and upper-bounds.
As can be seen in Figure~\ref{fig:noiseDistribution}(b,c), both datasets exhibit a long-tailed distribution, \ie, most identities have very few images. This phenomenon is especially obvious on the MegaFace~\cite{nech2017level} dataset since it uses automatically formed clusters for determining identities, therefore, the same identity may be distributed in different clusters.
Noises across all groups in MegaFace~\cite{nech2017level} are less in comparison to MS-Celeb-1M~\cite{guo2016ms}. 
However, we found that many images in the clean portion of MegaFace~\cite{nech2017level} are duplicated images. 
In Sec.~\ref{sec:damage_noise}, we will perform experiments on the MegaFace and MS-Celeb-1M datasets to quantify the effect of noise on the face recognition task.

\section{Building a Noise-Controlled Face Dataset}

As shown in the previous section, face recognition datasets that are more than a million scale typically have a noise ratio higher than 30\%. How about building a large scale noise controlled face dataset? It can be used to train better face recognition algorithms. More importantly, it can be used to further understand the relationship between noise and face recognition performance. To this end, we seek not only a cleaner and more diverse source to collect face data, but also an effective way to label the data. 

\begin{figure}[t]
\begin{center}
   \includegraphics[width=0.8\linewidth]{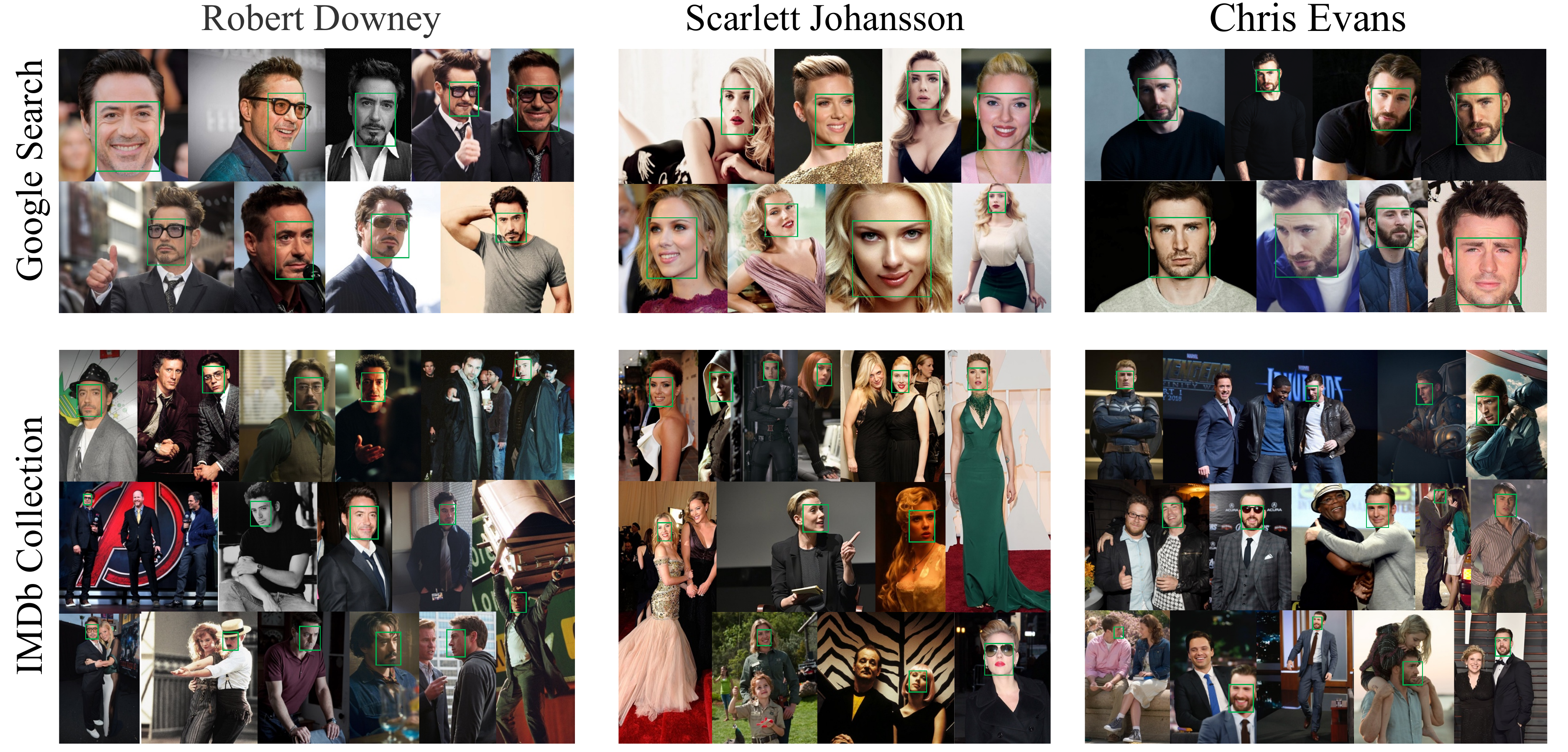}
\end{center}
\vskip -0.45cm
   \caption{The second row depicts the raw data from the IMDb website. As a comparison, we show the images of the same identity queried from the Google search engine in the first row.}
\label{fig:rawDataSample}
\end{figure}

\subsection{Celebrity Faces from IMDb} \label{sec:collecting}

Search engines are important sources from which we can quickly construct a large-scale dataset. The widely used ImageNet~\cite{deng2009imagenet} was built by querying images from Google Image. Most of the face recognition datasets were built in the same way (except MegaFace~\cite{nech2017level}). 
While querying from search engines offers the convenience of data collection, it also introduces data bias. Search engines usually operate in a high-precision regime~\cite{chen2014enriching}. 
Observing the queried images in Figure \ref{fig:rawDataSample}, they tend to have a simple background with sufficient illumination, and the subjects are often in a near frontal posture.  
These data, to a certain extent, are more restricted than those we could observe in reality, \eg, faces in videos (IJB-A~\cite{klare2015pushing} and YTF~\cite{wolf2011face}) and selfie photos (millions of distractors in MegaFace).
Another pitfall in crawling images from search engines is the low recall rate. We performed a simple analysis and found that on average the recall rate is only 40\% for the first 200 photos we query for a particular name.

In this study, we turn our data collection source to the IMDb website. IMDb is more structured. It includes a diverse range of photos under each celebrity's profile, including official photos, lifestyle photos, and movie snapshots. 
Movie snapshots, we believe, provide essential data samples for training a robust face recognition model. Those screenshots are rarely returned by querying a search engine. 
In addition, the recall rate is much higher (90\% on average) when we query a name on IMDb. This is much higher than 40\% from search engines.
The IMDb website lists about 300K celebrities who have official and gallery photos. By clawing IMDb dataset, we finally collected and cleaned 1.7M raw images from 59K celebrities.

\subsection{Data Distribution}\label{sec:dataDistribution}

\begin{figure*}[ht]
\begin{center}
   \includegraphics[width=\linewidth]{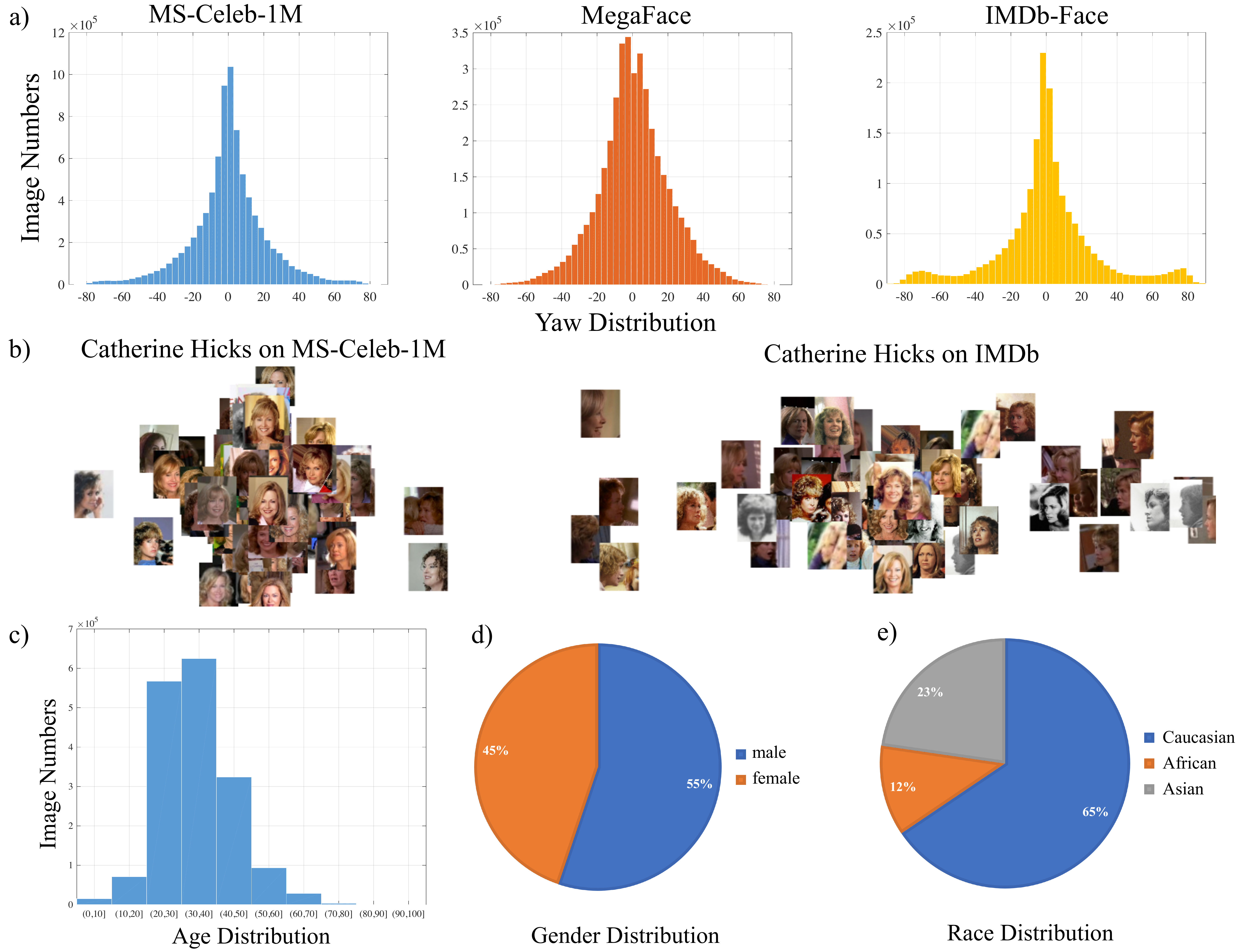}
\end{center}
\vskip -0.3cm
   \caption{a) Comparing the distribution of yaw angle of images in the proposed dataset against MS-Celeb-1M and MegaFace. b) A qualitative sample from the proposed IMDb-Face and MS-Celeb-1M. c) Age distribution of images in IMDb-Face. d) Gender distribution of identities in IMDb-Face. e) Race distribution of identities in IMDb-Face.}
\label{fig:yawDistribution}
\end{figure*}

Figure \ref{fig:yawDistribution}-a presents the distribution of yaw angle in our dataset compared with MS-Celeb-1M and MegaFace. Figures \ref{fig:yawDistribution}-c, -d and -e present the age, gender and race distributions.  As can be observed, images in IMDb-Face exhibit larger pose variations, and they also show diversity in age, gender and race.

\subsection{How Good can Human Label Identity?} \label{sec:userStudy}

The data downloaded from IMDb are noisy as multiple celebrities may co-exist on the same image. We still need to clean the dataset before it can be used for training. We take this opportunity to study how human annotators would clean a face data. The study will help us to identify the source of noise during annotation and design a better data cleaning strategy for the full dataset. 

\begin{figure}[t]
\begin{center}
   \includegraphics[width=\linewidth]{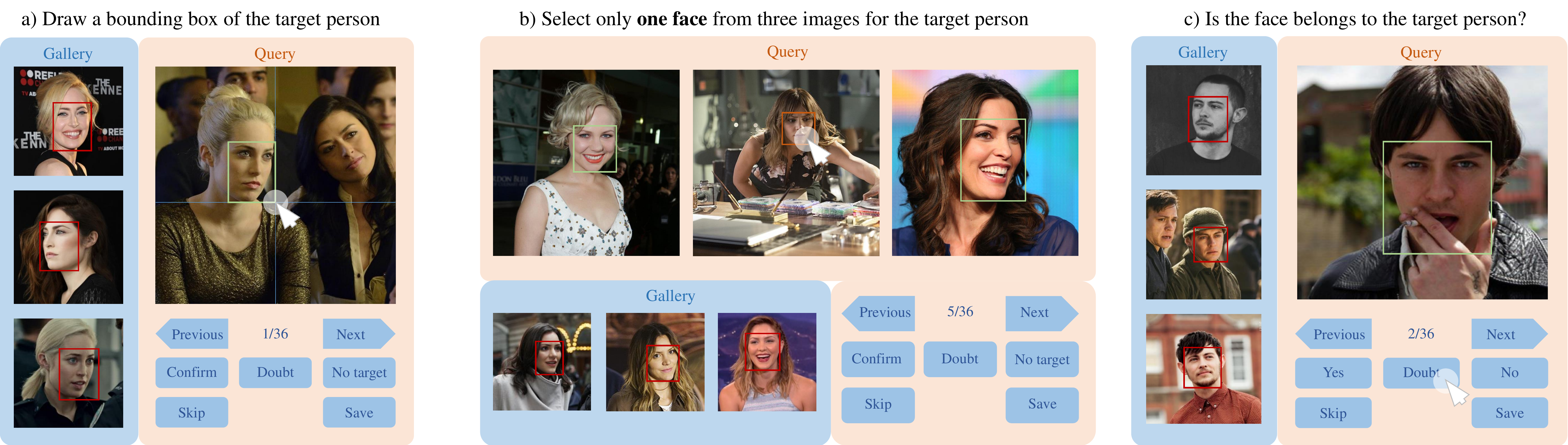}
\end{center}
\vskip -0.2cm
   \caption{Interfaces for user study: (a) Scheme I - volunteers were asked to draw a box on the target's face. (b) Scheme II - given three query faces, volunteers were asked to select the face that belongs to the target person. (c) Scheme III - volunteers were asked to select the face that belongs to the target.}
\label{fig:annotateInterface}
\end{figure}

For the purpose of the user study, we extract a small subset of 30 identities from the IMDb raw data. We carefully select three images with confirmed identity serving as gallery images. The remaining images of these 30 identities are treated as query images. To make the user study more challenging and statistically more meaningful, we inject 20\% outliers to the query set. 
Next, we prepare three annotation schemes as follows. The interface of each scheme is depicted in Figure~\ref{fig:annotateInterface}.

\noindent 
\textbf{Scheme I - Draw the box:} 
We present the target person to a volunteer by showing the three gallery faces. We then show a query image selected from the query set. The image may contain multiple persons. If the target appears in the query image, the volunteer is asked to draw a bounding box on the target. The volunteer can either confirm the selection or assign a `doubt' flag on the box if he/she is not confident about the choice. `No target' is selected when he/she cannot find the target person.

\noindent 
\textbf{Scheme II - Choose 1 in 3:} Similar to Scheme I, we present the target person to a volunteer by showing the gallery images. 
We then randomly sample three faces detected from the query set, from which the volunteer will select a single image as the target face. We ensure that all query faces have the same gender as the target person. Again, the volunteer can choose a `doubt' flag if he/she is not confident about the selection or choose `no target' at all.

\noindent
\textbf{Scheme III - Yes or No:} Binary query is perhaps be the most natural and popular way to clean a face recognition set. We first rank all faces based on their similarity to probe faces in the gallery, and then ask a volunteer to make a choice if each belongs to the target person. The volunteer is allowed to answer `doubt'.

\noindent
\textbf{Which scheme to choose?}: 
Before we can quantify the effectiveness of different schemes, we first need to generate the ground truth of these 30 identities. 
We use a `consensus' approach. Specifically, each of the aforementioned schemes was conducted on three different volunteers. We ensure that each query face was annotated nine times across the three schemes.
If four of the annotations consistently point to the same identity, we assign the query face to the targeted identity. With this ground truth, we can measure the effectiveness of each annotation scheme.

\begin{figure}[t]
\floatbox[{\capbeside\thisfloatsetup{capbesideposition={right,top},capbesidewidth=4cm}}]{figure}[\FBwidth]
{ \caption{A ROC comparison between three different annotating schemes; volunteers were allowed to select `doubt' so two data points can be obtained depending if we count doubt data as positive or negative.}\label{fig:humanROC}}
{\includegraphics[width=1.1\linewidth]{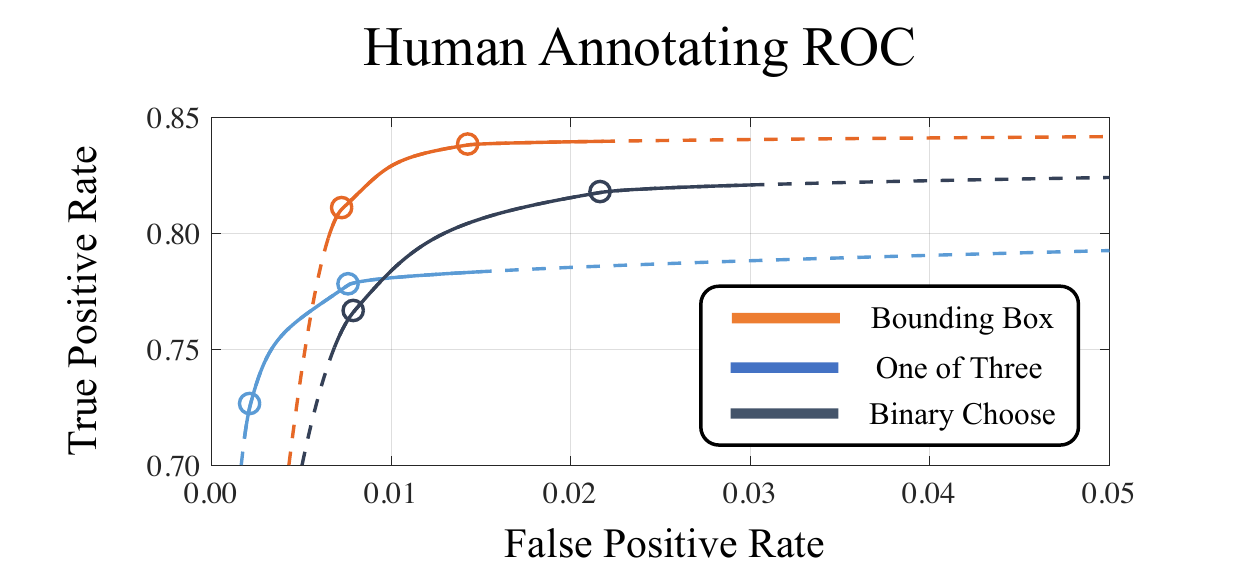}}
\end{figure}

Figure~\ref{fig:humanROC} shows the Receiver operating characteristic (ROC) curve of each of the three schemes\footnote{We should emphasize that the curves in Figure~\ref{fig:humanROC} are different from actual human's performance on verifying arbitrary face pairs. This is because in our study the faces from a query set are very likely to belong to the same person. The ROC thus represents human's accuracy on `verifying face pairs that likely belong to the same identity'.}.
Scheme I achieves the highest $F_1$ score. It recalls more than 90\% faces with under 10\% false positive samples. Finding a face and drawing a box seems to make annotators more focused on finding the right face.
Scheme II provides a high true positive rate when the false positive is low. The existence of distractors forces annotators to work harder to match the faces.
Scheme III yields the worse true positive rate when the false positive is low. This is not surprising since this task is much easier than Schemes I and II. The annotators tend to make mistakes given this relaxing task, especially after a prolonged annotation process.
We observe an interesting phenomenon: \textit{the longer a volunteer spends on annotating a sample, the more accurate the annotation is}. With full speed in one hour, each volunteer can draw 180-300 faces in Scheme I, or finish around 600 selections in Scheme II, or answer over 1000 binary questions in Scheme III.
We believe the most reliable way to clean a face recognition dataset is to leverage both Schemes I and II to achieve a high precision and recall. Limited by our budget, we only conducted Scheme I to clean the IMDb-Face dataset.

During the cleaning of the IMDb-Face, since multiple identities may co-exist on the same image, first we annotated gallery images to make sure the queried identity.
The gallery images come from the official gallery provided by the IMDb website, which most of these official gallery images contain the true identity. We ask volunteers to look through the 10 gallery images back and forth and draw bounding box of the face that occurs most frequently. 
Then, annotators label the rest of the queried images guided by the three largest labeled faces as galleries. For identities having fewer than three gallery images, their queried images may have too much noise. 
To save labor, we did not annotate their images.

It took 50 annotators one month to clean the IMDb-Face dataset. Finally, we obtained 1.7M clean facial images from 2M raw images. We believe that the cleaning is of high quality. We estimate the noise level of IMBb-Face as the product of approximated noise level in the IMDb raw data ($2.7 \pm 4.5$\%) and the false positive rate (8.7\%) of Scheme I. The noise level is controlled under 2\%. The quality of IMDb-Face is validated in our experiments.

\section{Experiments}
\label{experiment}

We divide our experiments into a few sections. First, we conduct ablation studies by simulating noise on our proposed dataset. The studies help us to observe the deterioration of performance in the presence of increasing noise, or when a fixed amount of clean data is diluted with noise. Second, we perform experiments on two existing datasets to further demonstrate the effect of noise. Third, we examine the effectiveness of our dataset by comparing it to other datasets with the same training condition. Finally, we compare the model trained on our dataset with other state-of-the-arts. Next, we describe the experimental setting.

\noindent
\textbf{Evaluation Metric:} 
We report rank-1 identification accuracy on the Megaface benchmark~\cite{kemelmacher2016megaface}. It is a very challenging task to evaluate the performance of face recognition methods at the million scale of distractors. The MegaFace benchmark consists of one gallery set and one probe set. The
gallery set contains more than 1 million images and the probe set consists of two existing datasets: Facescrub~\cite{ng2014data} and FGNet. We use Facescrub~\cite{ng2014data} as MegaFace probe dataset in our experiments. Verification performance of MegaFace (reported as TPR at FPR$=10^{-6}$) is included in the supplementary material due to page limit. We also test LFW~\cite{huang2007labeled} and YTF~\cite{wolf2011face} in Section \ref{sec:state-of-art}.

\noindent
\textbf{Architecture:} To better examine the effect of noise, we use the same architecture in all experiments. After a comparison among ResNet-50, ResNet-101 and Attention-56~\cite{Wang_2017_CVPR}, we finally choose Attention-56 that achieves a good balance between computation and accuracy. As a reference, the model converges on a database with 80 hours on an 8-GPU server with a batch-size of 256. The output of Attention-56 is a 256-dimensional feature for each input image. We use cosine similarity to compute scores between image pairs.

\noindent
\textbf{Pre-processing:} We cropped and aligned faces, then rigidly transferred them onto a mean shape. Then we resized the cropped image into $224\times256$, and subtracted them with the mean value in each RGB channel.

\noindent
\textbf{Loss:} We apply three losses: SoftMax~\cite{celebface}, Center Loss~\cite{wen2016discriminative} and A-Softmax~\cite{liu2017sphereface}. Our implementation is based on the public implementation of these losses: 

\noindent
\textit{Softmax:} Softmax loss is the most commonly used loss, either for model initialization or establishing a baseline.

\noindent
\textit{Center Loss:} Wen~\etal~\cite{wen2016discriminative} propose center loss, which minimizes the intra-class distance to enhance features' discriminative power. The authors jointly trained CNN with the center loss and the softmax loss. 

\noindent
\textit{A-Softmax:} Liu~\etal~\cite{liu2017sphereface} formulate A-Softmax to explicitly enforce the angle margin between different identities. The weight vector of each category was restricted on a hypersphere.

\subsection{Investigating the Effect of Noise on IMDb-Face}

The proposed IMDb-Face dataset enables us to investigate the effect of noise. 
There are two common types of noise in large-scale face recognition datasets:
1) \textit{label flips}: example has erroneously been given the label of another class within the dataset 
2) \textit{outliers}: image does not belong to any of the classes under consideration, but mistakenly has one of their labels. Sometimes even non-faces may be mistakenly included.
To simulate the first type of noise, we randomly perturb faces into incorrect categories. For the second type, we randomly replace faces in IMDb-Face with images from MegaFace.

\begin{figure}[ht]
\begin{center}
   \includegraphics[width=0.98\linewidth]{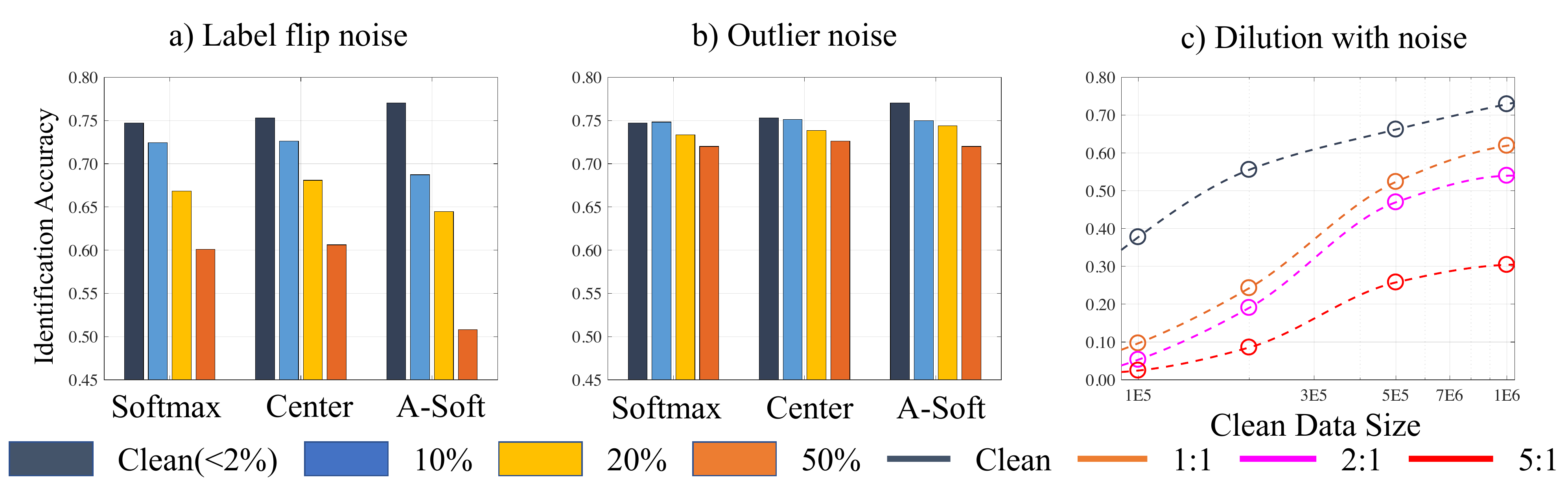}
\end{center}
\vskip -0.25cm
   \caption{1:1M rank-1 identification results on MegaFace benchmark: (a) introducing label flips to IMDb-Face, (b) introducing outliers to IMDb-Face, and (c) fixing the size of clean data and dilute it with different ratios of label flips.}
\label{fig:dilute}
\end{figure}

Here we perform two experiments: 
1) We gradually contaminate our dataset with different types of noise. We gradually increase the noise in our dataset by 10\%, 20\% and 50\%.
2) We fix the size of clean data and `dilute' it with label flips. 
We do not use ensemble models in these experiments.

Figure \ref{fig:dilute}(a) and (b) summarize the results of our first experiment. 1) Label flips severely deteriorate the performance of a model, more so than outliers. 2) A-Softmax, which used to achieve a better result on a clean dataset, becomes worse than Center loss and Softmax in the high-noise region. 3) Outliers seem to have a less abrupt effect on the performance across all losses, matching the observation in~\cite{krause2016unreasonable} and~\cite{rolnick2017deep}.

The second experiment was inspired by a recent work from Rolnick~\etal~\cite{rolnick2017deep}. They found that if a dataset contains sufficient clean data, a deep learning model can still be properly trained on it when the data is diluted by a large amount of noise. They show that a model can still achieve a feasible accuracy on CIFAR-10, even the ratio of noise to clean data is increased to $20:1$.
Can we transfer their conclusion to face recognition? Here we sample four subsets from IMDb-Face with $1E5$, $2E5$, $5E5$ and $1E6$ images. And we dilute them with an equal number, double, and five times of label flip noise. Figure \ref{fig:dilute}(c) shows that a large performance gap still exists against the completely clean baseline, even we maintain the same number of clean data.
We conjecture two reasons that cleanliness of data still plays a key role in face recognition: 1) current dataset, even it is clean, still far from sufficient to address the challenging face recognition problem thus noise matters. 2) Noise is more lethal on a 10,000-class problem than on a 10-class problem.

\subsection{The Effect of Noise on MegaFace and MS-Celeb-1M}
\label{sec:damage_noise}
To further demonstrate the effect of noise, we perform experiments on two public datasets: MegaFace and MS-Celeb-1M. 
In order to quantify the effect of noise on the face recognition, we sampled subsets from the two datasets and manually cleaned them. This provides us with a noisy sampled subset and a clean subset for each dataset.
For a fair comparison, the noisy subset was sampled to have the same distribution of image numbers to identities as the original dataset. Also, we control the scale of noisy subsets to make sure the scales for each clean subset are nearly the same. Because of the large size of the sampled subsets, we have chosen the third labeling method mentioned in Sec.~\ref{sec:userStudy}, which is the fastest.

\begin{table}[t]
\begin{center}
\caption{Noisy data vs. Clean data. The results are obtained from rank-1 identification test on the MegaFace benchmark~\cite{kemelmacher2016megaface}. Abbreviation MSV1 = MS-Celeb-1M(v1).}
\label{tab:test}
\footnotesize
\begin{tabular}{r|cc|ccc}
\hline
    \multirow{2}{*}{Dataset}  & \multirow{2}{*}{\#Iden.} & \multirow{2}{*}{\#Imgs.} &
       \multicolumn{3}{c}{MegaFace Rank-1(\%)} \\
       \cline{4-6}
       & & & Softmax & Center & A-softmax \\
       \hline
   MSV1-raw   & 96k   & 8.6M  & 71.70 & 73.82 & 73.99    \\
       -sampled             & 46k   & 3.7M  & 66.15 & 69.81 & 70.56    \\
       -clean               & 46k & 1.76M & 70.66 & 73.15 & 73.53   \\\hline
   MegaFace-raw          & 670k  & 4.7M  & 64.32 & 64.71 & 66.95   \\
            -sampled        & 270k  & 2.7M  & 59.68 & 62.55 & 63.12   \\
         -clean              & 270k  & 1.5M  & 62.86 & 67.64 & 68.88   \\\hline
   \end{tabular}
\end{center}
\end{table}

Three different losses, namely, SoftMax, Center Loss and A-Softmax, are respectively applied to the original datasets, sampled, and cleaned subsets. Table~\ref{tab:test} summarizes the results on the MegaFace recognition challenge~\cite{kemelmacher2016megaface}.
The effect of clean datasets is tremendous. By comparing the results between cleaned datasets and sampled datasets, the average improvement of accuracy is as large as $4.14\%$. The accuracies on clean subsets even surpass those on raw datasets, which are 4 times larger on average. The results suggest the effectiveness of reducing noise for large-scale datasets. As the mater of fact, the result of this experiment is part of our motivation to collect IMDb-Face dataset.

It is worth pointing out that recent metric learning based methods such as A-Softmax~\cite{liu2017sphereface} and Center-loss~\cite{wen2016discriminative} also benefit from learning on clean datasets, although they already perform much better than Softmax~\cite{celebface}. As shown in Table~\ref{tab:test}, the improvements of accuracy on MegaFace using A-Softmax and Center-loss are over $5\%$. The results suggest that reducing dataset noise is still helpful, especially when metric learning is performed. Reducing noisy samples could help an algorithm focuses more on hard examples learning, rather than picking up meaningless noises.

\subsection{Comparing IMDb-Face with other Face Datasets}

In the third experiment, we wish to show the competitiveness of IMDb-Face against several well-established face recognition training datasets including: 1) CelebFaces~\cite{sun2014deep,celebface}, 2) CASIA-WebFace~\cite{yi2014learning}, 3) MS-Celeb-1M(v1)~\cite{guo2016ms}, and 4) MegaFace~\cite{nech2017level}. 
The data size of the two latter datasets is a few times larger than the proposed IMDb-Face. Note that MS-Celeb-1M has a larger subset(v2), containing 900,000 identities. Limited by our computational resources we did not conduct experiments on it. We do not use ensemble models in this experiment.
Table \ref{tab:mainExperiment} summarizes the results of using different datasets as the training source across three losses. We observed that the proposed noise-controlled IMDb-Face dataset is competitive as a training source despite its smaller size, validating the effectiveness of the IMDb data source and the cleanliness of IMDb-Face.

\begin{table}[b]\small
\caption{Comparative results on using different face recognition datasets for training. Rank-1 identification accuracy on MegaFace benchmark is reported.}
\label{tab:mainExperiment}
\footnotesize
\setlength{\abovecaptionskip}{0pt}
\setlength{\belowcaptionskip}{-10pt}
\begin{center}
\begin{tabular}{r| c c|  c| c | c }
 \hline
    \multirow{2}{*}{Dataset}  &  \multirow{2}{*}{\#Iden.} & \multirow{2}{*}{\#Imgs.}  &
    \multicolumn{3}{c}{Rank-1 (\%)} \\
    \cline{4-6}
     & & & Softmax  & Center Loss & A-Softmax  \\
 \hline
 \hline
 CelebFaces  &10k & 0.20M  & 36.15  &42.54 &43.72\\
 \hline
 CASIA-WebFace  &10.5k & 0.49M  & 65.17  &68.09 &70.89\\
 \hline
 \multirow{1}{*}{MS-Celeb-1M(V1)} & 96k & 8.6M & 71.70 & 73.82 & 73.99 \\
 \hline
 \multirow{1}{*}{MegaFace}& 670k & 4.7M  & 64.32 & 64.71 & 66.95 \\
 \hline
 IMDbFace & 59k & 1.7M & \textbf{74.75} &\textbf{79.41}  & \textbf{84.06} \\
 \hline
\end{tabular}
\end{center}
\end{table}

\subsection{Comparisons with State-of-the-Arts} \label{sec:state-of-art}

We are interested to compare the performance of model trained on IMDb-Face with state-of-the-arts. Evaluation is conducted on MegaFace~\cite{kemelmacher2016megaface}, LFW~\cite{huang2007labeled}, and YTF~\cite{wolf2011face} following the standard protocol. 
For LFW~\cite{huang2007labeled} we compute equals error rate (EER). For YTF~\cite{wolf2011face} we report accuracy for recognition.
To highlight the effect of training data, we do not adopt model ensemble. The comparative results are shown in Table~\ref{tab:finalResult}. Our single model trained on IMDb-Face (A-Softmax$^\sharp$, IMDb-Face) achieves a state-of-the-art performance on LFW, MegaFace, and YTF against published methods. It is noteworthy that the performance of our final model is also comparable to a few private methods on MegaFace.

\begin{table}[t]\small
\footnotesize
\begin{center}
\caption{Comparisons with state-of-the-arts methods on LFW, MegaFace and YTF benchmarks.}
\begin{tabular}{r|cccc}
           Method, Dataset                      & LFW    & Mega(Ident.)  & YTF\\
           \hline
           Vocord-deep V3$^\dagger$, Private             & -      & \textbf{91.76}      & - \\
           YouTu Lab$^\dagger$, Private                   & -      & 83.29               & -  \\
           DeepSense V2$^\dagger$, Private                & -         & 81.23                   & - \\
           \hline
           \hline
           Marginal Loss$^\sharp$~\cite{deng2017marginal} MS-Celeb-1M        & 99.48     & 80.278   &95.98   \\
        SphereFace~\cite{liu2017sphereface},CASIA-WebFace                 & 99.42  & 75.77       & 95.00 \\
        Center Loss~\cite{wen2016discriminative},CASIA-WebFace                 & 99.28  & 65.24        & 94.90\\
        \hline
        \hline
        A-Softmax$^\sharp$, MS-Celeb-1M        &99.58    &    73.99      &    97.45    \\
        A-Softmax$^\sharp$, IMDb-Face         &\textbf{99.79}    &   \textbf{84.06}       &    \textbf{97.67} \\\hline
        \multicolumn{4}{l}{$\dagger$ Commercial, have not been published  } \\
        \multicolumn{4}{l}{$\sharp$ Single Model}\\
\end{tabular}
\vskip -0.3cm
\label{tab:finalResult}
\end{center}
\end{table}

\section{Conclusion}

Beyond existing efforts of developing sophisticated losses and CNN architectures, our study has investigated the problem of face recognition from the data perspective. 
Specifically, we developed an understanding of the source of label noise and its consequences. We also collected a new large-scale data from IMDb website, which is naturally a cleaner and wilder source than search engines. Through user studies, we have discovered an effective yet accurate way to clean our data. Extensive experiments have demonstrated that both data source and cleaning effectively improve the accuracy of face recognition. As a result of our study, we have presented a noise-controlled IMDb-Face dataset, and a state-of-the-art model trained on it.
A clean dataset is important as the face recognition community has been looking for large-scale clean datasets for two practical reasons: 
1) to better study the training performance of contemporary deep networks as a function of noise level in data. Without a clean dataset, one cannot induce controllable noise to support a systematic study.  
2) to benchmark large-scale automatic data cleaning methods. Although one can use the final performance of a deep network as a yardstick, this measure can be affected by many uncontrollable factors, \eg, network hyperparameters setting. A clean and large-scale dataset enables unbiased analysis.

%
%
%
\bibliographystyle{splncs04}
\bibliography{1201}
%




%
\end{document}